\documentclass[conference]{IEEEtran}

\ifCLASSINFOpdf
  \usepackage[pdftex]{graphicx}
  \graphicspath{ {./images/} }
\else
\fi
\usepackage{times}
\usepackage{amsmath}
\usepackage{booktabs}
\usepackage{setspace}
\usepackage{hyperref,cleveref}
\usepackage{listings}
\usepackage{esvect}
\usepackage{xspace}
\usepackage{tabularx}
\usepackage{comment}
\usepackage{afterpage}
\usepackage[T1]{fontenc}
\usepackage[utf8]{inputenc} 
\usepackage{url}            % simple URL typesetting
\usepackage{amsfonts}       % blackboard math symbols
\usepackage{nicefrac}       % compact symbols for 1/2, etc.
\usepackage{tikz}

\usepackage{microtype}
\usepackage{graphicx}
\usepackage{subfigure}
\usepackage{amssymb}
\usepackage{arydshln}
\usepackage{mathtools}
\usepackage{blkarray, bigstrut}
\usepackage[nolist]{acronym}

\usepackage{graphicx}
\usepackage{xspace}

\usepackage{latexsym}
\usepackage{comment}

\newcommand{\approach}{\textsc{Shallom}\xspace}
\newcommand{\triple}[3]{(\texttt{#1}, \texttt{#2}, \texttt{#3})}
\newcommand{\pair}[2]{(\texttt{#1}, \texttt{#2})}
\newcommand{\scoreFunc}{\psi}
\newcommand{\kg}{\ensuremath{\mathcal{G}}\xspace}
\newcommand{\entities}{\ensuremath{\mathcal{E}}}
\newcommand{\relations}{\ensuremath{\mathcal{R}}}

\newcommand{\classifierMatrix}{\ensuremath{\mathbf{W}}}
\newcommand{\hidden}{\ensuremath{\mathbf{H}}}
\newcommand{\biases}{\ensuremath{\mathbf{b}}}
\newcommand{\hiddenwidth}{\ensuremath{k}}
\newcommand{\embeddingDim}{\ensuremath{d}}
\newcommand{\concat}{\ensuremath{\Psi}}
\newcommand{\RealSpace}{\ensuremath{\mathbb{R}}}

\begin{acronym}[UML]
	\acro{AOS}{Agricultural Ontology Services}
	\acro{AGRIS}{Agricultural Science and Technology}
	\acro{API}{Application Programming Interface}
	\acro{AOS}{Agricultural Ontology Services}
	\acro{AGRIS}{Agricultural Science and Technology}
	\acro{API}{Application Programming Interface}
	\acro{A2KB}{Annotation to Knowledge Base}
	\acro{AI}{Artificial Intelligence}
	\acro{BPSO}{Binary Particle-Swarm Optimization}
	\acro{BPMLOD}{Best Practices for Multilingual Linked Open Data}
	\acro{BPSO}{Binary Particle-Swarm Optimization}
	\acro{BPMLOD}{Best Practices for Multilingual Linked Open Data}
	\acro{BFS}{Breadth-First-Search}
	\acro{BPE}{Byte Pair Encoding}
	\acro{BoW}{Bag-of-Words}
	\acro{CBD}{Concise Bounded Description}
	\acro{COG}{Content Oriented Guidelines}
	\acro{CSV}{Comma-Separated Values}
	\acro{CBMT}{Corpus-Based Machine Translation}
	\acro{CLIR}{Cross-Language Information Retrieval}
	\acro{DPSO}{Deterministic Particle-Swarm Optimization}
	\acro{DALY}{Disability Adjusted Life Year}
	\acro{DBMS}{Relational Database Management System}

	\acro{ER}{Entity Resolution}
	\acro{EM}{Expectation Maximization}
	\acro{EBMT}{Example-Based Machine Translation}
	\acro{EBNF}{Extended Backus--Naur Form}
	\acro{EL}{Entity Linking}
	\acro{FAO}{Food and Agriculture Organization of the United Nations}
	\acro{GIS}{Geographic Information Systems}
	\acro{GHO}{Global Health Observatory}
	\acro{GRU}{Gated recurrent unit}
	\acro{HDI}{Human Development Index}
	\acro{ICT}{Information and communication technologies}
	\acro{IFRS}{International Financial Reporting Standards}
	\acro{ICD}{International Classification of Diseases}
	\acro{IT}{Information Technology}
    \acro{KB}{Knowledge Base}
    \acro{KG}{Knowledge Graph}
    \acrodefplural{KG}{Knowledge Graphs}
    \acro{KGE}{Knowledge Graph Embeddings}
    \acro{KBSE}{Knowledge Base Semantic Embedding}
    \acro{KGC}{Knowledge Graph Completion}
	\acro{LR}  {Language Resource}
	\acro{LD}  {Linked Data}
	\acro{LLOD}  {Linguistic Linked Open Data}
	\acro{LIMES}{LInk discovery framework for MEtric Spaces}
	\acro{LS}  {Link Specifications}
	\acro{LDIF}{Linked Data Integration Framework}
	\acro{LGD} {LinkedGeoData}
	\acro{LOD} {Linked Open Data}
	\acro{LOV} {Linked Open Vocabularies}
	\acro{LSTM}{Long Short-Term Memories}
	\acro{MSE}{Mean Squared Error}
	\acro{MWE}{Multiword Expressions}
	\acro{MT}{Machine Translation}
	\acro{ML}{Machine Learning}
	\acro{MR}{Machine Reading}
	\acro{NIF}{Natural Language Processing Interchange Format}
	\acro{NIF4OGGD}{NLP Interchange Format for Open German Governmental Data}
	\acro{NLP}{Natural Language Processing}
	\acro{NER}{Named Entity Recognition}
	\acro{NMT}{Neural Machine Translation}
	\acro{NN}{Neural Network}
	\acrodefplural{NN}{Neural Networks}
	\acro{NLG}{Natural Language Generation}
	\acro{NED}{Named Entity Disambiguation}
	\acro{NERD}{Named Entity Recognition and Disambiguation}
	\acro{NL}{Natural Language}
	\acro{NIF}{NLP Interchange Format}
	\acro{NIF4OGGD}{NLP Interchange Format for Open German Governmental Data}
	\acro{NLP}{Natural Language Processing}
	\acro{NER}{Named Entity Recognition}
	\acro{NEL}{Named Entity Linking}
	\acro{NE}{Named Entity}
	\acro{NN}{Neural Network}
	\acro{NLI}{Natural Language Inference}
	\acro{OSM}{OpenStreetMap}
	\acro{OWL}{Web Ontology Language}
	\acro{OOV}{out-of-vocabulary}
	\acro{PFM}{Pseudo-F-Measures}
	\acro{PSO}{Particle-Swarm Optimization}
	\acro{PBSMT}{Phrase-Based Statistical Machine Translation}
	
	\acro{QA}{Question Answering}
	\acro{RDF}{Resource Description Framework}
	\acro{RBMT}{Rule-Based Machine Translation}
	\acro{RNN}{Recurrent Neural Network}
	\acro{ReLU}{rectified linear unit}
	\acro{RDFS}{RDF Schema}
	\acro{SKOS}{Simple Knowledge Organization System}
	\acro{SPARQL}{SPARQL Protocol and RDF Query Language}
	\acro{SRL}{Statistical Relational Learning}
	\acro{SWT}{Semantic Web Technologies}
	\acro{SW}{Semantic Web}
	\acro{SMT}{Statistical Machine Translation}
	\acro{SWMT}{Semantic Web Machine Translation}
	\acro{SKOS}{Simple Knowledge Organization System}
	\acro{SPARQL}{SPARQL Protocol and RDF Query Language}
	\acro{SRL}{Statistical Relational Learning}
	\acro{SF}{surface forms}
    \acro{TBMT} {Transfer-Based Machine Translation}
	\acro{UML}{Unified Modeling Language}
	\acro{USL}{Ukrainian Sign Language}
	\acro{URI}{Uniform Resource Identifier}
	\acro{WHO}{World Health Organization}
	\acro{WKT}{Well-Known Text}
	\acro{W3C}{World Wide Web Consortium}
	\acro{WSD}{Word Sense Disambiguation}
	\acro{WMT}{Workshop on Machine Translation}
    \acro{XML}{Extensible Markup Language}
	\acro{YPLL}{Years of Potential Life Lost}
	\acro{AOS}{Agricultural Ontology Services}
	\acro{AGRIS}{Agricultural Science and Technology}
	\acro{API}{Application Programming Interface}
	\acro{AOS}{Agricultural Ontology Services}
	\acro{AGRIS}{Agricultural Science and Technology}
	\acro{API}{Application Programming Interface}
	\acro{A2KB}{Annotation to Knowledge Base}
	\acro{BPSO}{Binary Particle-Swarm Optimization}
	\acro{BPMLOD}{Best Practices for Multilingual Linked Open Data}
	\acro{BPSO}{Binary Particle-Swarm Optimization}
	\acro{BPMLOD}{Best Practices for Multilingual Linked Open Data}
	\acro{BFS}{Breadth-First-Search}
	\acro{BPE}{Byte Pair Encoding}
	\acro{BoW}{Bag-of-Words}
	\acro{CBD}{Concise Bounded Description}
	\acro{COG}{Content Oriented Guidelines}
	\acro{CSV}{Comma-Separated Values}
	\acro{CBMT}{Corpus-Based Machine Translation}
	\acro{CLIR}{Cross-Language Information Retrieval}
	\acro{DPSO}{Deterministic Particle-Swarm Optimization}
	\acro{DALY}{Disability Adjusted Life Year}

	\acro{ER}{Entity Resolution}
	\acro{EM}{Expectation Maximization}
	\acro{EBMT}{Example-Based Machine Translation}
	\acro{EBNF}{Extended Backus--Naur Form}
	\acro{EL}{Entity Linking}
	\acro{FAO}{Food and Agriculture Organization of the United Nations}
	\acro{GIS}{Geographic Information Systems}
	\acro{GHO}{Global Health Observatory}
	\acro{GRU}{Gated recurrent unit}
	\acro{HDI}{Human Development Index}
	\acro{ICT}{Information and communication technologies}
	\acro{IFRS}{International Financial Reporting Standards}
	\acro{ICD}{International Classification of Diseases}
	\acro{IT}{Information Technology}
	\acro{IRI}{International Resource Identifier}
    \acro{KB}{Knowledge Base}
    \acro{KG}{Knowledge Graph}
    \acro{KGE}{Knowledge Graph Embeddings}
    \acro{KBSE}{Knowledge Base Semantic Embedding}
	\acro{LR}  {Language Resource}
	\acro{LD}  {Linked Data}
	\acro{LLOD}  {Linguistic Linked Open Data}
	\acro{LIMES}{LInk discovery framework for MEtric Spaces}
	\acro{LS}  {Link Specifications}
	\acro{LDIF}{Linked Data Integration Framework}
	\acro{LGD} {LinkedGeoData}
	\acro{LOD} {Linked Open Data}
	\acro{LSTM}{Long Short-Term Memories}
	\acro{MSE}{Mean Squared Error}
	\acro{MWE}{Multiword Expressions}
	\acro{MT}{Machine Translation}
	\acro{ML}{Machine Learning}
	\acro{MR}{Machine Reading}
	\acro{MOS}{Manchester OWL Syntax}
	\acro{NIF}{Natural Language Processing Interchange Format}
	\acro{NIF4OGGD}{NLP Interchange Format for Open German Governmental Data}
	\acro{NLP}{Natural Language Processing}
	\acro{NER}{Named Entity Recognition}
	\acro{NMT}{Neural Machine Translation}
	\acro{NN}{Neural Network}
	\acro{NLG}{Natural Language Generation}
	\acro{NED}{Named Entity Disambiguation}
	\acro{NERD}{Named Entity Recognition and Disambiguation}
	\acro{NL}{Natural Language}
	\acro{NIF}{NLP Interchange Format}
	\acro{NIF4OGGD}{NLP Interchange Format for Open German Governmental Data}
	\acro{NLP}{Natural Language Processing}
	\acro{NER}{Named Entity Recognition}
	\acro{NEL}{Named Entity Linking}
	\acro{NE}{Named Entity}
	\acro{NN}{Neural Network}
	\acro{NLI}{Natural Language Inference}
	\acro{OSM}{OpenStreetMap}
	\acro{OWL}{Web Ontology Language}
	\acro{OOV}{out-of-vocabulary}
	\acro{PFM}{Pseudo-F-Measures}
	\acro{PSO}{Particle-Swarm Optimization}
	\acro{PBSMT}{Phrase-Based Statistical Machine Translation}
	
	\acro{QA}{Question Answering}
	\acro{RDF}{Resource Description Framework}
	\acro{RBMT}{Rule-Based Machine Translation}
	\acro{RNN}{Recurrent Neural Network}
	\acro{ReLU}{rectified linear unit}
	\acro{SKOS}{Simple Knowledge Organization System}
	\acro{SPARQL}{SPARQL Protocol and RDF Query Language}
	\acro{SRL}{Statistical Relational Learning}
	\acro{SWT}{Semantic Web Technologies}
	\acro{SW}{Semantic Web}
	\acro{SMT}{Statistical Machine Translation}
	\acro{SWMT}{Semantic Web Machine Translation}
	\acro{SKOS}{Simple Knowledge Organization System}
	\acro{SPARQL}{SPARQL Protocol and RDF Query Language}
	\acro{SRL}{Statistical Relational Learning}
	\acro{SF}{surface forms}
	\acro{SVM}{Support Vector Machines}
    \acro{TBMT} {Transfer-Based Machine Translation}
	\acro{UML}{Unified Modeling Language}
	\acro{USL}{Ukrainian Sign Language}
	\acro{URI}{Uniform Resource Identifier}
	\acro{WHO}{World Health Organization}
	\acro{WKT}{Well-Known Text}
	\acro{W3C}{World Wide Web Consortium}
	\acro{WSD}{Word Sense Disambiguation}
	\acro{WWW}{World Wide Web}
    \acro{XML}{Extensible Markup Language}
	\acro{YPLL}{Years of Potential Life Lost}
	\acro{AOS}{Agricultural Ontology Services}
	\acro{AGRIS}{Agricultural Science and Technology}
	\acro{API}{Application Programming Interface}
	\acro{BPSO}{Binary Particle-Swarm Optimization}
	\acro{BPMLOD}{Best Practices for Multilingual Linked Open Data}
	\acro{CBD}{Concise Bounded Description}
	\acro{COG}{Content Oriented Guidelines}
	\acro{CSV}{Comma-Separated Values}
	\acro{CBMT}{Corpus-Based Machine Translation}
	\acro{CLIR}{Cross-Language Information Retrieval}
	\acro{DPSO}{Deterministic Particle-Swarm Optimization}
	\acro{DALY}{Disability Adjusted Life Year}

	\acro{ER}{Entity Resolution}
	\acro{EM}{Expectation Maximization}
	\acro{EBMT}{Example-Based Machine Translation}
	\acro{EBNF}{Extended Backus--Naur Form}
	\acro{EL}{Entity Linking}
	\acro{FAO}{Food and Agriculture Organization of the United Nations}
	\acro{GIS}{Geographic Information Systems}
	\acro{GHO}{Global Health Observatory}
	\acro{HDI}{Human Development Index}
	\acro{ICT}{Information and communication technologies}
    \acro{KB}{Knowledge Base}
    \acro{KBSE}{Knowledge Base Semantic Embedding}
	\acro{LR}  {Language Resource}
	\acro{LD}  {Linked Data}
	\acro{LLOD}  {Linguistic Linked Open Data}
	\acro{LIMES}{LInk discovery framework for MEtric Spaces}
	\acro{LS}  {Link Specifications}
	\acro{LDIF}{Linked Data Integration Framework}
	\acro{LGD} {LinkedGeoData}
	\acro{LOD} {Linked Open Data}
	\acro{MSE}{Mean Squared Error}
	\acro{MWE}{Multiword Expressions}
	\acro{MT}{Machine Translation}
	\acro{ML}{Machine Learning}
	\acro{NIF}{Natural Language Processing Interchange Format}
	\acro{NIF4OGGD}{NLP Interchange Format for Open German Governmental Data}
	\acro{NLP}{Natural Language Processing}
	\acro{NER}{Named Entity Recognition}
	\acro{NMT}{Neural Machine Translation}
	\acro{NN}{Neural Network}
	\acro{NLG}{Natural Language Generation}
	\acro{NED}{Named Entity Disambiguation}
	\acro{NERD}{Named Entity Recognition and Disambiguation}
	\acro{NL}{Natural Language}
	\acro{OSM}{OpenStreetMap}
	\acro{OWL}{Web Ontology Language}
	\acro{OOV}{out-of-vocabulary}
	\acro{PFM}{Pseudo-F-Measures}
	\acro{PSO}{Particle-Swarm Optimization}
	\acro{QA}{Question Answering}
	\acro{RDF}{Resource Description Framework}
	\acro{RBMT}{Ru\-le-Ba\-sed Ma\-chi\-ne Trans\-la\-tion}
	\acro{REG}{Referring Expression Generation}
	\acro{SKOS}{Simple Knowledge Organization System}
	\acro{SPARQL}{SPARQL Protocol and RDF Query Language}
	\acro{SRL}{Statistical Relational Learning}
	\acro{SWT}{Semantic Web Technologies}
	\acro{SW}{Semantic Web}
	\acro{SMT}{Statistical Machine Translation}
	\acro{SWMT}{Semantic Web Machine Translation}
    \acro{TBMT} {Transfer-Based Machine Translation}
	\acro{UML}{Unified Modeling Language}
	\acro{USL}{Ukrainian Sign Language}
	\acro{URL}{Uniform Resource Locator}
	\acro{WHO}{World Health Organization}
	\acro{WKT}{Well-Known Text}
	\acro{W3C}{World Wide Web Consortium}
	\acro{WSD}{Word Sense Disambiguation}
    \acro{XML}{Extensible Markup Language}
	\acro{YPLL}{Years of Potential Life Lost}
	\acro{AOS}{Agricultural Ontology Services}
	\acro{AGRIS}{Agricultural Science and Technology}
	\acro{API}{Application Programming Interface}
	\acro{AOS}{Agricultural Ontology Services}
	\acro{AGRIS}{Agricultural Science and Technology}
	\acro{API}{Application Programming Interface}
	\acro{A2KB}{Annotation to Knowledge Base}
	\acro{BPSO}{Binary Particle-Swarm Optimization}
	\acro{BPMLOD}{Best Practices for Multilingual Linked Open Data}
	\acro{BPSO}{Binary Particle-Swarm Optimization}
	\acro{BPMLOD}{Best Practices for Multilingual Linked Open Data}
	\acro{BFS}{Breadth-First-Search}
	\acro{BPE}{Byte Pair Encoding}
	\acro{BoW}{Bag-of-Words}
	\acro{CBD}{Concise Bounded Description}
	\acro{COG}{Content Oriented Guidelines}
	\acro{CSV}{Comma-Separated Values}
	\acro{CBMT}{Corpus-Based Machine Translation}
	\acro{CLIR}{Cross-Language Information Retrieval}
	\acro{DPSO}{Deterministic Particle-Swarm Optimization}
	\acro{DALY}{Disability Adjusted Life Year}

	\acro{ER}{Entity Resolution}
	\acro{EM}{Expectation Maximization}
	\acro{EBMT}{Example-Based Machine Translation}
	\acro{EBNF}{Extended Backus--Naur Form}
	\acro{EL}{Entity Linking}
	\acro{FAO}{Food and Agriculture Organization of the United Nations}
	\acro{GIS}{Geographic Information Systems}
	\acro{GHO}{Global Health Observatory}
	\acro{GRU}{Gated recurrent unit}
	\acro{HDI}{Human Development Index}
	\acro{ICT}{Information and communication technologies}
	\acro{IFRS}{International Financial Reporting Standards}
	\acro{ICD}{International Classification of Diseases}
	\acro{IT}{Information Technology}
    \acro{KB}{Knowledge Base}
    \acro{KG}{Knowledge Graph}
    \acro{KGE}{Knowledge Graph Embeddings}
    \acro{KBSE}{Knowledge Base Semantic Embedding}
	\acro{LR}  {Language Resource}
	\acro{LD}  {Linked Data}
	\acro{LLOD}  {Linguistic Linked Open Data}
	\acro{LIMES}{LInk discovery framework for MEtric Spaces}
	\acro{LS}  {Link Specifications}
	\acro{LDIF}{Linked Data Integration Framework}
	\acro{LGD} {LinkedGeoData}
	\acro{LOD} {Linked Open Data}
	\acro{LSTM}{Long Short-Term Memories}
	\acro{MSE}{Mean Squared Error}
	\acro{MWE}{Multiword Expressions}
	\acro{MT}{Machine Translation}
	\acro{ML}{Machine Learning}
	\acro{MR}{Machine Reading}
	\acro{NIF}{Natural Language Processing Interchange Format}
	\acro{NIF4OGGD}{NLP Interchange Format for Open German Governmental Data}
	\acro{NLP}{Natural Language Processing}
	\acro{NER}{Named Entity Recognition}
	\acro{NMT}{Neural Machine Translation}
	\acro{NN}{Neural Network}
	\acro{NLG}{Natural Language Generation}
	\acro{NED}{Named Entity Disambiguation}
	\acro{NERD}{Named Entity Recognition and Disambiguation}
	\acro{NL}{Natural Language}
	\acro{NIF}{NLP Interchange Format}
	\acro{NIF4OGGD}{NLP Interchange Format for Open German Governmental Data}
	\acro{NLP}{Natural Language Processing}
	\acro{NER}{Named Entity Recognition}
	\acro{NEL}{Named Entity Linking}
	\acro{NE}{Named Entity}
	\acro{NN}{Neural Network}
	\acro{NLI}{Natural Language Inference}
	\acro{OSM}{OpenStreetMap}
	\acro{OWL}{Web Ontology Language}
	\acro{OOV}{out-of-vocabulary}
	\acro{PFM}{Pseudo-F-Measures}
	\acro{PSO}{Particle-Swarm Optimization}
	\acro{PBSMT}{Phrase-Based Statistical Machine Translation}
	
	\acro{QA}{Question Answering}
	\acro{RDF}{Resource Description Framework}
	\acro{RBMT}{Rule-Based Machine Translation}
	\acro{RNN}{Recurrent Neural Network}
	\acro{ReLU}{rectified linear unit}
	\acro{SKOS}{Simple Knowledge Organization System}
	\acro{SPARQL}{SPARQL Protocol and RDF Query Language}
	\acro{SRL}{Statistical Relational Learning}
	\acro{SWT}{Semantic Web Technologies}
	\acro{SW}{Semantic Web}
	\acro{SMT}{Statistical Machine Translation}
	\acro{SWMT}{Semantic Web Machine Translation}
	\acro{SKOS}{Simple Knowledge Organization System}
	\acro{SPARQL}{SPARQL Protocol and RDF Query Language}
	\acro{SRL}{Statistical Relational Learning}
	\acro{SF}{surface forms}
	\acro{SVM}{Support Vector Machines}
    \acro{TBMT} {Transfer-Based Machine Translation}
	\acro{UML}{Unified Modeling Language}
	\acro{USL}{Ukrainian Sign Language}
	\acro{URI}{Uniform Resource Identifier}
	\acro{WHO}{World Health Organization}
	\acro{WKT}{Well-Known Text}
	\acro{W3C}{World Wide Web Consortium}
	\acro{WSD}{Word Sense Disambiguation}
    \acro{XML}{Extensible Markup Language}
	\acro{YPLL}{Years of Potential Life Lost}
\end{acronym}  

\begin{document}

\title{A shallow neural model for relation prediction}

\author{\IEEEauthorblockN{Caglar Demir}
\IEEEauthorblockA{Data Science Group\\ 
Paderborn University \\
North Rhine-Westphalia, Germany \\
Email: first.lastname@upb.de}
\and
\IEEEauthorblockN{Diego Moussallem}
\IEEEauthorblockA{Data Science Group\\ 
Paderborn University \\
North Rhine-Westphalia, Germany \\
Email: first.lastname@upb.de}
\and
\IEEEauthorblockN{Axel-Cyrille Ngonga Ngomo}
\IEEEauthorblockA{Data Science Group\\ 
Paderborn University \\
North Rhine-Westphalia, Germany \\
Email: first.lastname@upb.de}}

\maketitle

\begin{abstract}
Knowledge graph completion refers to predicting missing triples. Most approaches achieve this goal by predicting entities, given an entity and a relation. We predict missing triples via the relation prediction. To this end, we frame the relation prediction problem as a multi-label classification problem and propose a shallow neural model (\approach) that accurately infers missing relations from entities. \approach is analogous to C-BOW as both approaches predict a central token (\texttt{p}) given surrounding tokens (\pair{s}{o}). Our experiments indicate that \approach outperforms state-of-the-art approaches on the FB15K-237 and WN18RR
with margins of up to $3\%$ and $8\%$ (absolute), respectively,  while requiring a maximum training time of 8 minutes on these datasets. We ensure the reproducibility of our results by providing an open-source implementation including training and evaluation scripts at {\url{https://github.com/dice-group/Shallom}.}
\end{abstract}

\section{Introduction}
%intro - completed

\acp{KG} represent structured collections of facts describing the world in the form of typed relationships between entities~\cite{hogan2020knowledge}. These collections of facts have been applied to diverse tasks, including machine translation and collaborative filtering~\cite{kge_survey,moussallem2019augmenting}. However, most \acp{KG} on the Web suffer from incompleteness~\cite{nickel2015review}. For instance, the birth place of $71\%$ of the persons in Freebase and $66\%$ of the persons in DBpedia is not to be found in the respective \acp{KG}. In addition, more than $58\%$ of the scientists in DBpedia are not linked to the predicate that describes what they are known for~\cite{krompass2015type}. The identification of such missing information is called knowledge graph completion~\cite{shi2017proje} that is addressed in predicting missing entities or relations.  
Knowledge graph embedding approaches have been particularly successful at the knowledge graph completion task, among many others~\cite{dettmers2018convolutional,kge_survey,demir2020physical}.

We investigate the use of a shallow \acp{NN} for predicting missing triples. The motivation thereof lies in the following consideration: Several early works have shown that \acp{NN} (even with a single hidden layer) are universal approximators~\cite{cybenko1989approximation}. This means that shallow \acp{NN} with numerous non-polynomial activation functions approximate any continuous function on a complex domain. However, these theorems do not impose a constraint on the number of units in the hidden layer~\cite{goodfellow2016deep}. In addition, deep \acp{NN} seem to perform better than shallow \acp{NN} when the target function is expected to be a hierarchical composition of functions~\cite{poggio2017and}. Still, training deep \acp{NN} requires more extensive hyperparameter optimization than training shallow \acp{NN} to alleviate the overfitting problem and the choice of initialization technique plays a more important role for deep \ac{NN} in their applications~\cite{goodfellow2016deep}. Moreover, deep \acp{NN} necessitate more computational resources, have higher energy consumption, and consequently lead to substantially higher $CO_2$ emissions~\cite{strubell2019energy}. The essay of the hardware lottery~\cite{hooker2020hardware} highlighted the impact of available hardware system in determining which research ideas succeed (and
fail). It is therein emphasized how  the hardware lottery can delay research progress by casting successful ideas as failures. Importantly, findings of Ruffinelli et al.~\cite{ruffinelli2019you} have shown that the relative performance differences between various KGE approaches often shrinks and sometimes even reverses when compared to prior results provided that approaches are optimized properly. With this consideration, we propose a shallow neural model, \approach, for relation prediction that relies on
two affine transformations. By virtue of this architecture, \approach is analogous to C-BOW~\cite{mikolov2013efficient}, as both approaches predict a central token (\texttt{p}) given surrounding tokens \pair{s}{o}. 

We evaluate our approach against many state-of-the-art approaches on the WN18, WN18RR, FB15K, FB15K-237, and YAGO3-10 benchmark datasets. Overall, our results suggest that \approach outperforms the state-of-the-art in terms of Hits at N (Hits@N) and 
has a more efficient runtime. In particular, \approach yields state-of-the-art performance with a training time of under ten minutes on a knowledge graph containing more than $10^6$ triples.
    
\section{Preliminaries and Notation}
\label{sec:preliminaries}

\subsection{Knowledge Graph and Completion}
\label{subsec:rdf}

Let \entities\ and \relations\ represent the set of entities and relations, respectively. Then, a \ac{KG} $\kg= \{\triple{s}{p}{o}  \in \entities \times \relations \times \entities\}$ can be formalised as a set of triples where each triple contains two entities $\texttt{s},\texttt{o} \in \entities$ and a relation $\texttt{r} \in \relations$. \ac{KGC} refers to predicting missing triples on a given \kg. Most approaches learn a scoring function $\scoreFunc$ that is often formalised as $\scoreFunc:\entities \times \relations \times \entities \mapsto \mathbb{R}$~\cite{dettmers2018convolutional}. 
In contrast, the scoring function of approaches solely addressing the relation prediction task is often defined as $\scoreFunc:\entities \times \entities \mapsto  \mathbb{R}^{|\relations|}$ ~\cite{onuki2019relation}. Both formalizations allow computing a score for any triple \triple{s}{p}{o} either directly (i.e., by computing $\scoreFunc\triple{s}{p}{o}$) in the case of the entity prediction or indirectly (i.e., by looking up the value for \texttt{p} in $\scoreFunc\pair{s}{o}$) for the relation prediction. Ergo, \ac{KGC} approaches differ primarily in their scoring function $\scoreFunc$ while sharing the same goal: given an \triple{s}{p}{o}, its score is expected to be proportional to the likelihood of such a triple being contained $\triple{s}{p}{o} \in \kg$~\cite{dettmers2018convolutional}. To learn such function, most \ac{KGC} approaches generate corrupted/negative examples~\cite{bordes2013translating,nickel2015review}. In this setting, each $\triple{s}{p}{o} \in \kg$ is considered as a positive example, whilst all $\triple{x}{y}{z} \not\in \kg$ with $\texttt{x,z} \in \entities$ and $\texttt{y} \in \relations$ are considered to be candidates for negative examples~\cite{nickel2015review}.
% To learn such function, entity prediction approaches often employ the idea of generating negative triples, initially proposed in~\cite{bordes2013translating}, where each $\triple{s}{p}{o} \in \kg$ is considered as a positive example, whilst all $\triple{x}{y}{z} \not\in \kg$ with $\texttt{x,z} \in \entities$ and $\texttt{y} \in \relations$ are considered to be candidates for negative examples~\cite{nickel2015review}.
Ergo, such approaches presuppose that the absence of a relationship between two entities implies that the corresponding triple is \textbf{false} if such triple is sampled as a corrupted triple otherwise \textbf{unknown}. Such a schema creates a \textbf{trichotomy} (positive, negative and unknown triples) and disregards the \emph{open world assumption}, which suggests that non-existing triples are to be interpreted as unknown, not false~\cite{nickel2015review}. \approach complies with the open world assumption since the learning problem is formulated as a multi-label classification problem
where a dichotomy between triples is created.
% $\triple{s}{p}{o}\subseteq \entities \times \relations \times \entities$ is considered to be either known or unknown.

\section{\approach}
\label{sec:approach}
In this section, we formally elucidate \approach that is defined as
\begin{equation}
    \label{scoring_func_architecture}
    \scoreFunc(s,o)=  \sigma\Big(\classifierMatrix \cdot \text{ReLU} \big( \hidden \cdot \concat( s,o) + \biases_1 \big)+\biases_2 \Big),
\end{equation}
where $\concat(s,o) \in \mathbb{R}^{2\embeddingDim}$, $\hidden \in \mathbb{R}^{ \hiddenwidth \times 2\embeddingDim }$, $\classifierMatrix \in \mathbb{R}^{ |\relations| \times \hiddenwidth}$, $\biases_1 \in \mathbb{R}^\hiddenwidth$, and $\biases_2 \in \mathbb{R}^{|\relations|}$. $\sigma(\cdot)$, ReLU$(\cdot)$ and $\concat(\cdot,\cdot)$ denote the sigmoid, the rectified linear unit and the vector concatenation functions, respectively. Given \pair{s}{o}, $\concat(s,o)$ returns concatenated embeddings of $\pair{s}{o}$. Thereafter, we perform two affine transformations with the ReLU and the sigmoid function to obtain predicted probabilities for relation ($\hat{\mathbf{y}}\in \RealSpace^{|\relations|}$). Finally, the incurred loss is computed by the binary cross-entropy function:
% $\mathcal{L}(\mathbf{y}, \hat{\mathbf{y}})$ where $\hat{\mathbf{y}}$ is the vector of predicted probabilities and $\mathbf{y}$ is a binary vector indicating multi labels.
\begin{equation}
    \label{loss}
  \mathcal{L}(\mathbf{y}, \hat{\mathbf{y}}) = - \sum_i ^{|\relations|} \Big( (\mathbf{y}_i \cdot \log( \hat{\mathbf{y}}_i ) ) + (1-\mathbf{y}_i) \cdot \log(1- \hat{\mathbf{y}}_i) \Big)
\end{equation}
where $\hat{\mathbf{y}}$ is the vector of predicted probabilities and $\mathbf{y}$ is a binary vector of indicating multi labels.

\begin{figure*}[htb]
    \centering
    % .tikz presumably cause "! LaTeX Error: File `tkz-graph.sty' not found. " at uploading to arvix.
    %\input{shallom.tikz}
    \includegraphics[scale=.3]{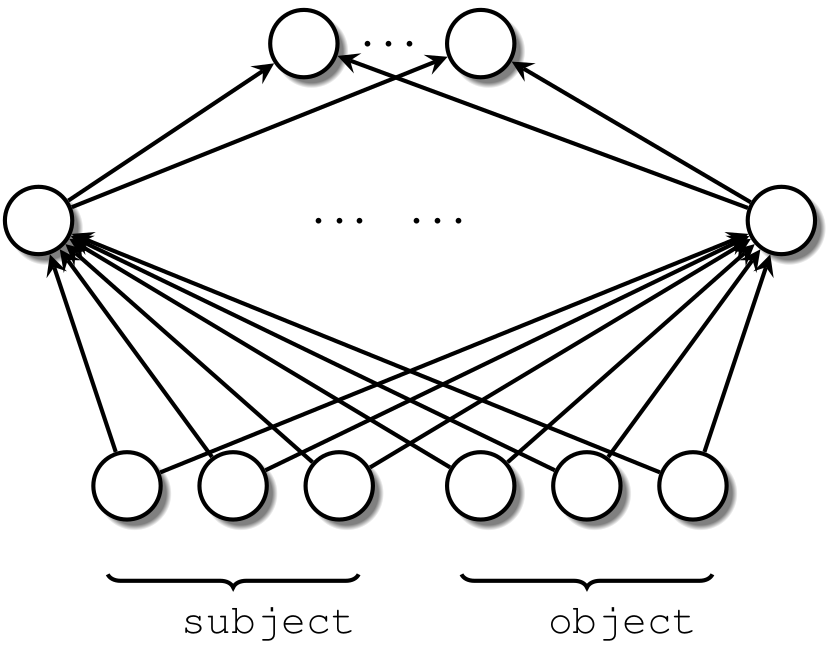}
    \caption{Visualization of \approach.}
    \label{fig:shallom}
\end{figure*}

~\Cref{fig:shallom} shows the architecture of \approach. To obtain a composite representation of \pair{s}{o}, we concatenate embeddings of entities as opposed to averaging them, since averaging embeddings loses the order of the input (as in the standard bag-of-words representation~\cite{le2014distributed}). Retaining order of embeddings avoids possible loss of information. As concatenation does not consider any interaction between the latent features, the first affine transformation is applied with the ReLU activation function. Thereafter, the second affine transformation is applied with the sigmoid function
to generate probabilities for relations.

\section{Experiments}
\label{sec:experiments}

We compared \approach against many state-of-the-art approaches and~\texttt{Uniform Random Classifier} (URC) in the relation prediction task on benchmark datasets \cite{dettmers2018convolutional}.

\subsection{Evaluation Protocol}
\label{subsubsec:evaluation metrics}
% Completed.

We applied Hits@N to evaluate the prediction performances. Given a test triple $\triple{s}{p}{o} \in \kg^{\text{Test}}$, we computed $|\relations|$ number of scores and obtained $I_{\text{relation}}=[(\texttt{p'},score): score=\scoreFunc(s,p',o) \wedge p' \in \relations]$. Then, we sorted the $I_{\text{relation}}$ in descending order of assigned scores and we computed Hits@N as follows:
  \begin{equation}
      \text{Hits@N = } \displaystyle{\frac{1}{|\kg^{\text{Test}}|} \sum_{\triple{s}{p}{o} \in \kg^{\text{Test}}} f( I_{\text{relation}}, \texttt{p} , N) },
   \label{metrics}
\end{equation}
where $f$ returns $1$ if \texttt{p} is contained in the top $N$ ranked tuples, otherwise $0$~\cite{shi2017proje}. To evaluate runtime performances, we measured the elapsed runtime during the training phase. Ergo, we ignored the elapsed time during the data preprocessing since the training setup for \approach is done on the fly while some approaches, including RDF2Vec, require additional computations such as applying the random walk technique. All approaches were trained four times on datasets. The reported runtimes (RT) of approaches are in seconds and the mean of the last three runs.

\subsection{Hyperparameter Optimization}
\label{subsec:hyperparameter_optimization}
% Completed.
We selected the hyperparameters of \approach via grid search according to the Hits@1 on the validation set of each dataset. The hyperparameter ranges for the grid search were set as follows: embedding size $\embeddingDim=[30, 50, 100, 200]$, $\text{epochs}=[30,50,100]$, the width of the hidden layer $\hiddenwidth=[.5\embeddingDim,\embeddingDim,3\embeddingDim]$, $\text{batch size}=[256,1000]$, $\text{dropout rate}=[.0, .2, .5]$ and $L_2\text{-normalizer}=[.0,.1]$. Initially, we used the default hyperparameters for all competing approaches provided in~\cite{trouillon2017knowledge}. However, RESCAL, ComplEx, CP and DistMult did not terminate within three hours of computation. The long runtimes are corroborated by~\cite{trouillon2017knowledge}. We hence optimized the hyperparameters of RESCAL, CP, TransE, DistMult and ComplEx via a grid search according to the Hits$@1$ on the validation set of each dataset. The hyperparameter ranges for the grid search were as follows: $\text{epochs}=[100,200]$, negative ratio per valid triple $=[1,5,10,50]$, and batch size $=[256, 512,|\kg^{\text{Train}}|/100]$. We omitted $\embeddingDim$, regularization term and learning rate from grid-search and used the parameter settings provided in ~\cite{trouillon2017knowledge}. We selected the hyperparameters of RDF2Vec via grid search according to the Hits$@1$ on the validation set of each dataset. The hyperparameter ranges of $\text{RDF2Vec}$ for the grid search were set as follows: embedding size $\embeddingDim=[50, 100]$, $\text{epochs}=[100]$, $\text{number of negatives for W2V}=[25,100]$ and $\text{random walk depth}=[3,5,7]$. After the embedding vectors are generated, we train the same scoring function defined in~\Cref{scoring_func_architecture} (look-up operation performed on RDF2Vec embeddings), by following the same optimization schema as our approach.

\section{Results}
\label{sec:results}
~\Cref{results_on_WN18_FB15},~\Cref{results_on_WN18RR_FB15K237} 
and~\Cref{results_on_yago_dbpedia} report the HitsN relation prediction results on the five benchmark datasets. Overall, \approach outperforms many state-of-the-art approaches while maintaining a superior runtime performance. 
% On the FB15K and WN18,~\approach outperforms most approaches. 
The slightly superior (.018 absolute) performance of ProjE on the FB15K comes with the cost of more than $3$ hours of computation.~\approach is significantly more time-efficient; it requires only 8 minutes, on average, a commodity computer. Since we could not reproduce the reported relation prediction results~\cite{shi2017proje}, we could neither re-evaluate ProjE on FB15K nor include it on the other benchmark datasets. Approaches perform significantly better on WN18 than on FB15K. This may stem from the fact that WN18 contains (1) significantly fewer relations and (2) entity pairs having multiple relations than FB15K. More specifically, FB15K and WN18 datasets contain $63.856$ and $277$ number entity pairs, respectively, that occurred with multiple relations in the training splits. 
\begin{table}[t]
    \small
    \caption{Hits@1 relation prediction results on FB15K and WN18.
    Results are taken from corresponding papers.}
    \centering
    \begin{tabular}{lcc}
    \hline\bf{Method} & FB15K & WN18\\\hline
    TransE~\cite{shi2017proje} & .651 &.736\\
    TransR~\cite{xie2016representation} & .702 & .713\\
    ProjE-listwise~\cite{shi2017proje} & \textbf{.758}&-\\ 
    PTransE (ADD, len-2 path)~\cite{shi2017proje}& .695 &-\\ 
    DKLR(CNN)~\cite{xie2016representation} & .698 &-\\
    TKRL (RHE)~\cite{xie2016representationtkrl} & .711 &-\\
    RDFDNN~\cite{onuki2017predicting} & .691 & .770 \\
    KGML~\cite{onuki2019relation} & .725  & \textbf{.975}\\
    SSP~\cite{xiao2017ssp} & .709 &-\\
    \midrule
    $\approach$ & .734 & .970\\
    \bottomrule
\label{results_on_WN18_FB15}
\end{tabular}
\end{table}
% Results on WN18RR and FB15k-237
\Cref{results_on_WN18RR_FB15K237} shows that~\approach outperforms all state-of-the-art approaches on the WN18RR and FB15K-237 datasets while maintaining an overall superior runtime performance. Note that the RT solely denotes the elapse training runtime (see \cref{subsubsec:evaluation metrics} for details.) Initially, we trained RESCAL, TransE, ComplEx, CP and DistMult with hyperparameters provided in~\cite{trouillon2016complex}. However, models other than TransE did not terminate within $3$ hours of computation. Consequently, we selected the hyperparameters of approaches via grid search as explained in~\Cref{subsec:hyperparameter_optimization}. TransE and DistMult yield a surprisingly better performance on WN18RR and FB15K-237 than on WN18 and FB15K. This may stem from (1) the hyperparameter optimization and (2) the fact that fewer numbers of entity pairs have multiple relations on training and testing datasets. The hyperparameters of TransE were not optimized in~\cite{shi2017proje,onuki2017predicting,onuki2019relation} where the Hit@1 performances of TransE were taken. CP performed poorly on the WN18RR due to the small number of relations as observed in~\cite{trouillon2017knowledge}. During the training phase, the batch size was set to $32$ in KGML and RDFDNN~\cite{onuki2019relation,onuki2017predicting}. Although training models with a small-batch regime seemed to alleviate a possible degradation in the generalization performances of models~\cite{keskar2016large}, it came with the cost of increased runtime. By virtue of being a shallow \ac{NN}, the error propagation was computationally more efficient in \approach than KGML. Importantly, KGML and RDFDNN do not optimize the width of the hidden layers. Conversely, we optimized the width of \approach, as per the suggestion in~\cite{nguyen2018neural}---that optimizing the width of the network has an impact on the generalization performance. RDFDNN erroneously assumes one-to-one mapping between entity pairs to relations and possibly suffers from the hyperbolic tangent saturation as the hyperbolic tangent is applied in the hidden layer~\cite{krizhevsky2012imagenet}. RDF2Vec outperforms RESCAL, TransE, CP and DistMult w.r.t. Hits@3 and Hits@5 on WN18RR. 
\begin{table*}[t]
\small
\caption{The mean of Hits@N relation prediction and runtime results on WN18RR and FB15K-237.}
    \centering	
        \begin{tabular}{lccccccccccc}
    \toprule
	 & \multicolumn{4}{c}{{ \bf WN18RR}} & &  \multicolumn{4}{c}{{ \bf FB15K-237}}                  \\
	\cmidrule{2-5}
	\cmidrule{7-12} 
	& & & 
	\multicolumn{1}{c}{Hits} 
	& & & & & 
	\multicolumn{1}{c}{Hits}  \\
	\cmidrule{3-5}     	
	\cmidrule{8-12}         
	            & RT & @1 & @3 & @5 & & RT & @1 & @3 & @5  \\
	\midrule
	RESCAL      & 1860$\pm$6 & .331 & .529 & .734   & & 5160$\pm$4 & .115 & .327 & .456 \\
	TransE      & 960$\pm$11 & .507 & .761 & .864   & & 540$\pm$10 & .774 & .899 & .918 \\
	ComplEx     & 2160$\pm$15 & .515 & .652 & .758   & & 5880$\pm$30 & .153 & .300 & .378 \\
	CP          & 840$\pm$15  & .332 & .518 & .659   & & 8040$\pm$39 & .467 & .609&  .675\\
	DistMult    & 780$\pm$13 & .497  & .677 & .799    & & 1140$\pm$8& .092 & .176 & .428 \\
	\midrule
	KGML        & 840$\pm$15  & .868 & .954 & .975   & & 1080$\pm$10 &.921 & .960 & .976 \\
	RDFDNN      & 540$\pm$8  & .819 & .967 & .985   & & 720$\pm$10 & .913 & .934 & .953 \\
	$\text{RDF2Vec}_{\text{Skip-Gram}}$          &  310$\pm$5 & .534 & .815 & .940   & &  482$\pm$6& .518  & .600 & .677 \\
	$\text{RDF2Vec}_{\text{CBOW}}$          &  337$\pm$10 & .451 & .785 & .932   & & 472$\pm$8 & .522  & .608 & .687 \\
	URC          &   & .095 & .265 & .446   & &  & .003  & .013 & .020 \\
	\midrule
	\approach    &  610$\pm$13 & \textbf{.874}  & \textbf{.982} & \textbf{.995}   & &  404$\pm$8 & \textbf{.948}  & \textbf{.993} & \textbf{.997} \\
	\bottomrule
\end{tabular}
    \label{results_on_WN18RR_FB15K237}
\end{table*}

\begin{table*}[t]
\small
\caption{The mean of Hits@N relation prediction and runtime results on YAGO3-10.}
    \centering	
        \begin{tabular}{lccccccccccc}
    \toprule
	 & \multicolumn{4}{c}{{ \bf YAGO3-10}}\\
	\cmidrule{2-5}
	& & & 
	\multicolumn{1}{c}{Hits} \\
	\cmidrule{3-5}     	
    & RT & @1 & @3 & @5\\
	$\text{RDF2Vec}_{\text{Skip-Gram}}$   & 593$\pm$11 &  .487 & .796  & .875\\
	$\text{RDF2Vec}_{\text{CBOW}}$        & 625$\pm$12 &  .491 & .803  & .873\\
	\midrule
	\approach         & 562$\pm$19  & \textbf{.630}  & \textbf{.983} & \textbf{.996} \\
		\bottomrule
\end{tabular}
    \label{results_on_yago_dbpedia}
\end{table*}
To confirm the performance of \approach, we compared it with some of the best approaches in terms of runtime requirement and Hits@1 on a large benchmark dataset. ~\Cref{results_on_yago_dbpedia} shows that \approach reaches close to $1.0$ Hits@5 and requires less than $10$ minutes on the YAGO3-10. We could not evaluate KGML on YAGO3-10 due to its high memory consumption requiring more than 16 GB RAM.

% \section{Discussion}
% \label{discussion}
The superior performance of \approach stems from: (1) it being a shallow neural model, (2) optimizing the width of the hidden layer, (3) the task and evaluation measures used. % Point:1
By virtue of being a shallow \ac{NN}, \approach requires only $562$ seconds to train on $|\kg| >10^6$ on a commodity computer.
% Point:2
\acp{NN} are required to be wide enough (larger than the input dimension) to learn disconnected decision regions~\cite{nguyen2018neural}. % It has also been empirically shown that classes of wide networks exist that cannot be accomplished by narrow networks.
% Choosing the width of the network plays an important role. Compared to KGML and RDFDNN, we optimize the width of the \approach.
% Point:3
Lastly, given the example \pair{Obama}{Hawaii}, \approach assigns high scores for \texttt{BirthPlace} and low scores for \texttt{SpouseOf}. This stems from the fact that input \kg does not involve triples such as \pair{SpouseOf}{Hawaii}, while it involves many triples \pair{BirthPlace}{Hawaii}. \approach assigns presumably a high score
\triple{Obama}{BirthPlace}{Paderborn}
although such a triple is not contained in \kg. Since the test splits of the benchmark datasets do not involve such false triples, the Hit$@N$ metric quantifies merely the performances of the relation prediction approaches on the valid triples. Ergo, the idea of corrupted triples is not necessary for relation prediction as each entity pair found in the test split is linked with a relation. 

\section{Related work}
\label{sec:related work}
A wide range of works have investigated the \ac{KGC} problem~\cite{nickel2015review,ji2020survey}. DistMult~\cite{yang2015embedding} can be seen as an efficient extension of RESCAL with a diagonal matrix per relation. ComplEx~\cite{trouillon2016complex} extends DistMult into a complex vector space. RDFDNN~\cite{onuki2017predicting} considers the relation prediction task as a multi-class classification problem. Embeddings of entities are learned disjointly. Relations are predicted through the softmax function. KGML~\cite{onuki2019relation} implements a multi-layer neural model for relation prediction. Experimental results show that KGML outperforms TransE, TransR, PTransE and RDFDNN in the relation prediction task. After an embedding layer, KGML concatenates embeddings of subject and object with the element-wise product of embeddings. This is followed by two inner product layers with fixed decreasing width. 
\approach differs from KGML by: (1) concatenating embeddings of entities without including a multiplication step, (2) having solely one inner product layer and omitting dynamic weighted binary cross entropy loss. RDF2Vec~\cite{ristoski2016rdf2vec} employs Word2Vec~\cite{mikolov2013efficient} for unsupervised feature extraction from sequences of words, and adapts them to RDF graphs.

\section{Conclusion}
\label{sec:conclusion}
% completed
We presented a shallow neural model effectively predicts missing triples without disregarding the \emph{open world assumption}.
\approach retains a linear space complexity in the number of entities. Experiments showed that training \approach on benchmark datasets is completed within a few minutes. This is an important result, as it means that our approach can be applied on large knowledge graphs without requiring high-performance hardware. This also implies that winning the hardware lottery is not necessary to tackle the link prediction problem~\cite{hooker2020hardware}. In future work, we plan to investigate extending \approach into temporal knowledge graphs and learning complex-valued valued embeddings via \approach \cite{trabelsi2018deep}, \cite{demir2020convolutional}.

\section*{Acknowledgment}
This work has been supported by the BMWi-funded project RAKI (01MD19012D) as well as the BMBF-funded project DAIKIRI (01IS19085B).
\bibliographystyle{IEEEtran}
\bibliography{references}
% that's all folks
\end{document}